\def\year{2019}\relax
\pdfoutput=1
\documentclass[letterpaper]{article} 
\usepackage{aaai19}  
\usepackage{times}  
\usepackage{helvet}  
\usepackage{courier}  
\usepackage{graphicx}  
\usepackage{booktabs}
\usepackage{multirow}
\usepackage{url}
\usepackage{float}
\usepackage{CJK}
\usepackage{amsmath}
\begin{document}
\begin{CJK}{UTF8}{gbsn}

\title{Neural Entity Reasoner for Global Consistency in Named Entity Recognition}




  \author{{Xiaoxiao Yin$^\dagger$, Daqi Zheng$^\ddagger$, Zhengdong Lu$^\ddagger$, RuiFang Liu$^\dagger$}\\
  $^\dagger$Beijing University of Posts and Telecommunications \{ygyyxx110, lrf\}@bupt.edu.cn\\
  $^\ddagger$DeeplyCurious.ai \{da, luz@deeplycurious.ai\} 
 }


\maketitle

\begin{abstract}
We propose Neural Entity Reasoner (NE-Reasoner), a framework to introduce global consistency of recognized entities into Neural Reasoner over Named Entity Recognition (NER) task.
Given an input sentence, the NE-Reasoner layer can infer over multiple entities to increase the global consistency of output labels, which then be transfered into entities for the input of next layer. 
NE-Reasoner inherits and develops some features from Neural Reasoner 1) a symbolic memory, allowing it to exchange entities between layers. 2) the specific interaction-pooling mechanism, allowing it to connect each local word to multiple global entities, and 3) the deep architecture, allowing it to bootstrap the recognized entity set from coarse to fine.
Like human beings, NE-Reasoner is able to accommodate ambiguous words and Name Entities that rarely or never met before.
Despite the symbolic information the model introduced, NE-Reasoner can still be trained effectively in an end-to-end manner via parameter sharing strategy.
NE-Reasoner can outperform conventional NER models in most cases on both English and Chinese NER datasets. 
For example, it achieves state-of-art on CoNLL-2003 English NER dataset.

\end{abstract}

\section{Introduction}

As a task to find and identify the named entities (NEs) such as person, location etc. in a text, Named Entity Recognition (NER) is considered as a basic and low-level problem of text understanding. NER is often solved as a sequence labeling problem (\cite{Collobert:11}, \cite{Strubell:17}, \cite{Huang:15}, \cite{Ma:16}, \cite{Chiu:16}, \cite{Lample:16}).

Current methods make decisions through the calculation results of word representations in the input sequence, so they actually rely on local linguistic features learned from the training dataset to recognize NEs.  Most recent research focus on introducing more information to leverage these linguistic features better (\cite{Peters:17}, \cite{Liu:18}, \cite{Zhang:18}).  

So actually, existing methods can not figure out what a word really means in an article, they can only try to make a decision in a local scope independently, so it's difficult to keep the consistency of these decisions.  To our humans, it's kind like we only read a short paragraph of a sentence one time, and every time we don't remember what we read before, in this situation, we can not understand quite well at all, so as the neural networks.

However, the expression of human language is variable and complex, and the recognition of named entity is actually to understand what role a word play in the entire article, not only in a local scope of a sentence.  Therefore, when there are ambiguous words or NEs that rarely be met before, existing models relied on local linguistic features cannot get enough information to make the right decisions.

In our human reading process, we read a word not just based on the knowledge we had before, we can look at the entire article, remember all local information while reading, and then connect and integrate local knowledge to understand the meaning of words by combining the entire article.  Even in some extreme cases, we may need to read until the end of the article to understand what a word really means.

Inspired by the human mind, we introduce a high-level "inference" mechanism to NER task and proposed NE-Reasoner. NE-Reasoner has a multi-layered architecture, each layer completes NER independently, and the NER results of each layer will be stored through a candidate pool as the reference for the next layer.  Because of this design, when making local decisions, the model can "see" and refer to relevant decisions elsewhere in the same article, so as to make wiser decisions. 

The candidate pool can be viewed as an external symbolic memory which is composed of entities, so it's different from existing Memory Augmented Neural Networks (MANNs) (\cite{Graves:14}, \cite{Weston:15}).  The reference from candidate pool is implemented by a special neural network, which is actually a multi-facts inference model(\cite{Peng:15}), then NE-Reasoner can rely on this to keep global consistency through reasoning.  

We implemented our model on both English and Chinese NER datasets, the result shows that our model can make better decisions in many cases compared to the first-layer NER results, and achieved state-of-art on CoNLL-2003 English NER dataset.

\section{Related works}

As a basic task in NLP, NER has received a lot of attention and research, and the neural networks work pretty well on this problem.  \cite{Collobert:11} proposed a CNN-CRF structure, it could be viewed as an encoder-decoder model, which used CNNs as encoder and CRF as decoder.  Since then, many works are based on this architecture, \cite{Strubell:17}  presents an application of \cite{Yu:16} for NER, which replace the CNN from \cite{Collobert:11} with dilated CNN to get improvement.   More recent works use LSTM as encoder since RNNs (LSTM, GRU) based models perform better and more natural in sequential problems, \cite{Huang:15} used bidirectional LSTM as encoder, and the Bi-LSTM-CRF model achieved state-of-art on many datasets, even be viewed as a standard method for sequence labeling.  \cite{Ma:16} and \cite{Chiu:16} proposed a hierarchical structure which used an additional CNN to represent character-level features,  while \cite{Lample:16} used LSTM instead, this kind of character encoder can extract features inside words and get better representations.  Not only the encoder, there are also lots of work on the decoder,  \cite{Mesnil:13} and \cite{Nguyen:16} used RNN as tag decoder, the RNN decoder takes the predict result of last time step as a feedback and input it into the RNN unit to learn the tags transfer rules form it, it turned out to outperform CRF decoder on some NER datasets.  

As an addition to the calculation of encoder and decoder, how to get better representations of words and better leverage word information has always received research attention, and most current methods focus on introducing more information.  \cite{Zhang:18} used lattice LSTM to get better leveraging explicit word information in Chinese.  \cite{Rei:17}and \cite{Peters:17} introduced pre-trained language model in sequence labeling and leverage external information resource.  \cite{Liu:18} presented a task-aware neural language model to extract abundant knowledge hidden in raw texts to empower the sequence labeling task and then achieved state-of-art on many datasets.  Meanwhile how to leverage extra information from the results is also a popular approach in sequential problems.  \cite{Xia:17} used an additional second-pass decoder to refine the result from the first-pass decoder in sequence generate tasks since the second time has global information about what the sequence to be generated might be.  In sequence labeling task, this kind information from the predict results is also very useful, \cite{Xu:17} used a second-pass decoder to augment the first-pass performance by replacing the named entities recognized in the first-pass with their predicted types in the second-pass, in this work, each layer of our model completes NER independently, and we actually use the entities information from the results of the previous layer as external information to help in the next layer.

No matter how great the representations we get, existing methods based on local decisions cannot achieve real reasoning. Memorizing is a direct method to save and leverage information across long distance, \cite{Graves:14} and \cite{Weston:15} proposed a framework that use an external memory to augment neural networks，MANNs based on this framework can be designed flexibly to handle different tasks.  The memory can help to get better ability to remember facts from the past, \cite{Sukhbaatar:15} and \cite{Kumar:16} presented end-to-end memory networks that can be used to solve question answering problem.  \cite{Santoro:16} showed MANNs can learn to bind data representations to their appropriate labels and achieved great performance on meta-learning.  \cite{Wang:16} introduced this kind of external memory to neural machine translation (NMT), the memory can provide a more flexible way to leverage the information of source sentence.  \cite{Tang:16} used memory to store symbolic information and let the model learn to leverage external knowledge. 

However, most existing end-to-end MANNs leverage memory in a distributed way and keep it in global, so it's difficult to be trained and used on sequence labeling tasks.  To our knowledge, we are the first to introduce external symbolic memory to sequence labeling task.  

\section{Overview of NE-Reasoner}
\label{overview}
\begin{figure}[h]
\centerline{\includegraphics[scale=0.35]{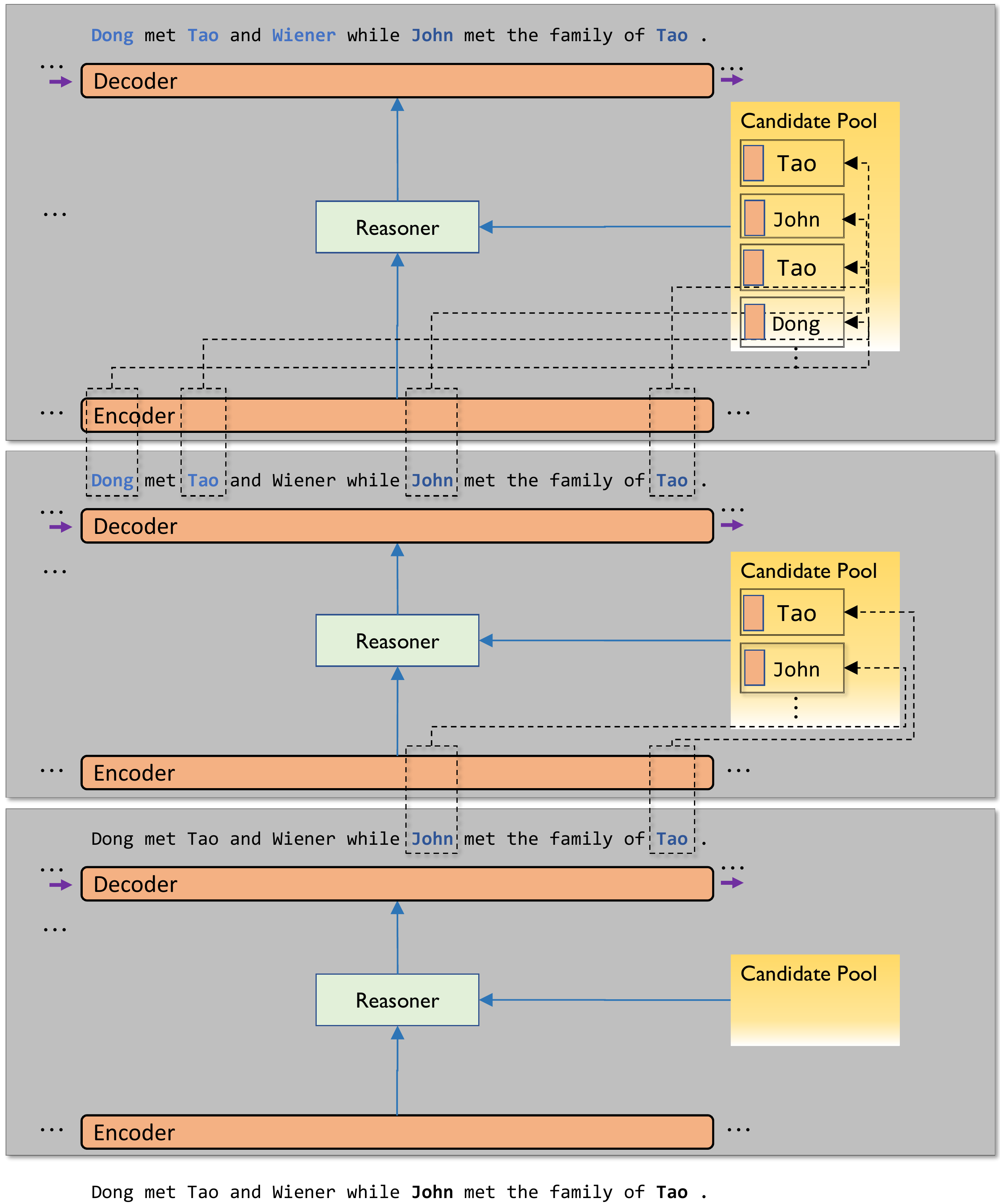}}
\caption{The overall architecture of our model. \label{overall_fig}}
\end{figure} 

NE-Reasoner model has a multi-layered architecture, each layer is an encoder-decoder structure NER model that can complete NER independently.  The input sequence is encoded into a sequence of vector representations which are generated from the word representations and their context information, then the decoder of each layer can rely on this to give predict results independently.  The predict results label out which words are entities, so the encoding representations of these words contain the information that could be relied on to make these decisions, in another word, we can find out the entities representations from the predict results.  

Meanwhile, our model keeps a candidate pool throughout the entire NER process, which consists of entities information recognized already.  The model completes NER layer by layer, so it can "see" all decisions made past, then each layer can take reference from it through the reasoner, and update this candidate pool from the predict results to help the next layer to keep global consistency and get better results.  

The candidate pool is like a cache memory which is only updated after the NER process of each layer.  The update operation relies on the predict labels, which are kind of symbolic information.  We don't need to keep the candidate pool continuous because it won't change its content in each layer, it's just like an external source of information consists of entities information independently, and the reasoner is actually a multi-facts inference model to get an answer as the reference from it.  Actually, although the information in the candidate pool is vector representations, but the operations and leveraging are a more symbolic way in neural networks framework.  

For example, as shown in Figure \ref{overall_fig}, the input sequence "Dong met Tao and Wiener while John met the family of Tao" has some words which are difficult to make decisions for existing methods, so in the first layer, the candidate pool is empty because there are no entities recognized before, then the model gives out predict results just like other existing models, it can recognize "John" cause it's a regular, common person name that appears a lot in the training dataset, and also the last "Tao" could be labeled out because the context pattern "met the family of", but rest of entities don't have enough and strong signals to be recognized correctly.

Then we can store the information of both entities into the candidate pool through the results, so in the second layer, the model can make an inference from it by the reasoner.  The model can know that the word before "met" could be a person name from the information of "John", and also know "Tao" is a person name, so it can dig out "Dong" and the first "Tao" by these two inferences so as to keep consistency, then update the candidate pool.  Just like the operation in the second layer, the model can recognize "Wiener" by reasoner in the third layer and complete the entire NER process.  

\section{Model}

In this section, we will introduce the overall architecture of our model and then explain the details of individual components.

\subsection{Overall Architecture}

As we shown in \ref{overview}, a NE-Reaoner layer can be described more specifically as follow:

\begin{itemize}
\item Get word representations of input sequence ${\textsl X=\{\mathbf x_{1},\mathbf x_{2},\mathbf x_{3},...,\mathbf x_{n}}\}$ by word embedding and charecter level CNN.
\item Encode the input sequence into a sequence of vector representations ${\textsl H=\{\mathbf h_{1},\mathbf h_{2},\mathbf h_{3},...,\mathbf h_{n}}\}$ that contain information of each word and their context.  
\item Use encoding information ${\textsl H}$ and entities information in candidate pool to make an inference by the reasoner and get the reference ${\textsl S}$.
\item Feed encoding information ${\textsl H}$ and reference ${\textsl S}$ into the decoder and get predict results ${\mathbf y=\{y_{1},y_{2},y_{3},...,y_{n}}\}$.
\item Update the candidate pool through the predict results ${\mathbf y}$.
\end{itemize}

Therefore a NE-Reasoner layer could be viewed as a regular encoder-decoder framework based NER model, which can take extra information by reasoner, as we know, the encoder-decoder framework could have many variants, and NE-Reasoner could be implemented on all of it.

In this work, our model uses Bi-LSTM model as the encoder, LSTM model as the decoder and a character level CNN.  The candidate pool is a simple list that consists of the encoding information of entities which can be selected by the predict results, and it can contain all entities recognized in the entire article or mini-batch because it is built on the entire results.  

The decoders and encoders in all layers can share parameters to avoid growing of parameters and make the model easy to train as an end-to-end model, so the only difference between each layer is the difference of the candidate pool and the entities they can refer to. 

\subsection{Encoder and Decoder}

Without the reference from the reasoner and the candidate pool, the encoder and decoder of each layer are a regular CNN-Bi-LSTM-LSTM NER model, the input sequence ${\textsl X=\{\mathbf x_{1},\mathbf x_{2},\mathbf x_{3},...,\mathbf x_{n}}\}$ is fed into the encoder which is a bidirectional RNNs model, in this work we use the Long Short-Term Memory network (LSTM) proposed by \cite{Hochreiter:97}. 

\subsubsection{Bidirectional LSTM}

LSTM is a variant RNN designed to solve the problems of gradient vanishing and learning long-term dependencies.  Formally, at time $t$, the memory $c_{t}$ and the hidden state $h_{t}$ of the basic LSTM unit are updated with the following equations: 

\begin{equation}
\begin{split}
\label{E1}
\begin{bmatrix}
\mathbf{\tilde{c}_{t}}\\ 
\mathbf{o_{t}}\\ 
\mathbf{i_{t}}\\ 
\mathbf{f_{t}}
\end{bmatrix}
&=
\begin{bmatrix}
tanh\\ 
\sigma \\ 
\sigma \\ 
\sigma 
\end{bmatrix}
(\mathbf{W}\begin{bmatrix}
\mathbf{x_{t}}\\ 
\mathbf{h_{t-1}}
\end{bmatrix}
+\mathbf{b}),\\
\mathbf{c_{t}} &= \mathbf{\tilde{c}_{t}}\odot \mathbf{i_{t}}+\mathbf{c_{t-1}}\odot \mathbf{f_{t}},\\
\mathbf{h_{t}} &= \mathbf{o_{t}}\odot tanh(\mathbf{c_{t}})
\end{split}
\end{equation}

where $\odot$ is the element-wise product, $\sigma$ is the sigmoid function and $x_{t}$ is the input vector at time $t$.  $\mathbf{o_{t}}$, $\mathbf{o_{t}}$, $\mathbf{o_{t}}$ denote the input, forget and output gates at time step $t$ respectively.

LSTM takes only information before the current input word, but the context information behind could also be crucial in sequential tasks.  To capture the context information from both past and future, the Bi-LSTM uses another LSTM to encode the sequence from end to start, then we can get the hidden states ${\textsl H=\{\mathbf h_{1},\mathbf h_{2},\mathbf h_{3},...,\mathbf h_{n}}\}$ as follows:

\begin{equation}
\label{E2}
\mathbf h_{t}=\textup{Bi-LSTM}(\mathbf x_{t})=[\mathbf h_{t}^{f}, \mathbf h_{t}^{b}]
\end{equation}

where $h_{t}^{f}$ is the hidden state of forward LSTM, $h_{t}^{b}$ is the hidden state of backward LSTM.

\subsubsection{Decoder}

The decoding layer is an uni-direction LSTM, which uses ${\textsl H}$ to predict tags ${\mathbf y^{i}=\{y_{1}^{i},y_{2}^{i},y_{3}^{i},...,y_{n}^{i}}\}$, and also takes the predict of last time step as input as follows, where $i$ denotes the $i$-th layer.  
\begin{equation}
\begin{split}
\label{E3}
\mathbf p_{t}^{i}&=Softmax(\textup{LSTM}(\mathbf h_{t},\mathbf p_{t-1}^{i}))\\
y_{t}^{i}&=argmax(\mathbf p_{t}^{i})
\end{split}
\end{equation}

\begin{figure}[h]
\centerline{\includegraphics[scale=0.35]{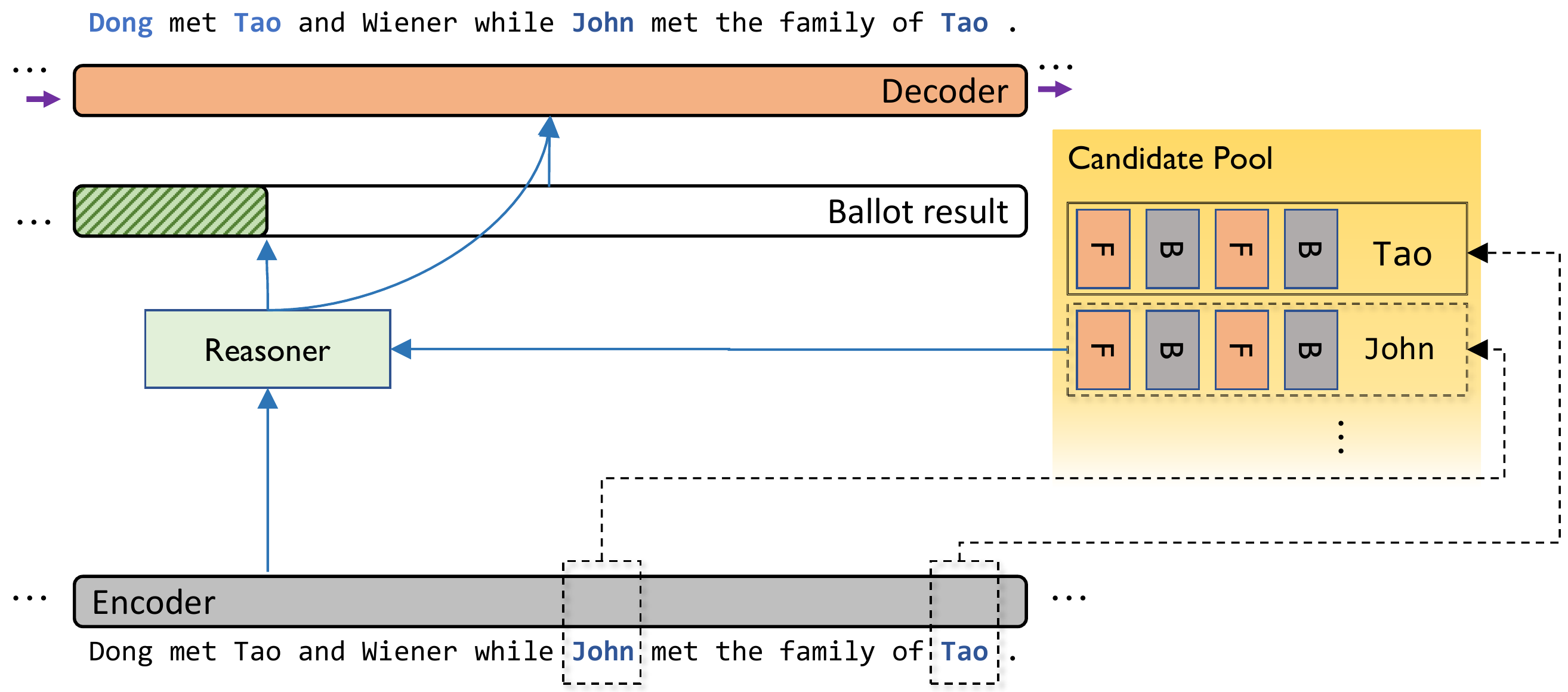}}
\caption{The details of how to generate the document memory from the first-pass predict results. \label{candidate}}
\end{figure} 

\subsection{Candidate Pool}
\label{memory}
We can get a pretty good predict ${\mathbf y^{i}}$ since we use the Bi-LSTM-LSTM model as our baseline which has been used on many prior works.  We adopt the BMEOS (Begin, Middle, End, Other, Single) tagging scheme, so we can tell where is the beginning or end of every entity from ${\mathbf y^{i}}$, and then use this boundary information to organize and form the cache memory of document.  

Since the model relies on the local language features to make decisions, we can consider how to store entities information more reasonable and effective from this. Obviously, in our human reading process, we treat an entity as an independent and indivisible object which is composed of a couple of words, so the pattern that an entity appears could be described like this:

[Forward Context] [Entity] [Backward Context]

To let the model learn to take reference from other entities, the model needs to store entities in this kind of pattern.  Like we mentioned above, the encoding information of each entity contains the information to decide it's label, so we can figure out how the model make decisions from this.  

The encoder in encoding layer is a combination of a forward LSTM and a backward LSTM, so we can divide $\mathbf h_{t}$ into $\mathbf h_{t}^{f}$ and $\mathbf h_{t}^{b}$ like \ref{E2}

where $\mathbf h_{t}^{f}$ is computed by the forward LSTM, and $\mathbf h_{t}^{b}$ is computed by the backward LSTM, the hidden state of an uni-LSTM is a computation of the input representations before the time $t$, so the information of $\mathbf h_{t}^{f}$ and $\mathbf h_{t}^{b}$ have different meaning clearly.  The decoder decide which word is beginning and which word is end through $\mathbf h_{t}^{f}$ and $\mathbf h_{t}^{b}$, at the beginning, $\mathbf h_{t}^{f}$ is used to capture the forward context information, $\mathbf h_{t}^{b}$ is used to capture the entity information, and at the end, $\mathbf h_{t}^{f}$ is used to capture entity information from another direction, $\mathbf h_{t}^{b}$ is used to capture the backward context information.

Therefore, the encoding information of the beginning and end of an entity can totally represent this entity, which contains the information about what this entity looks like and how we understand it from the context.  We can locate the boundary of entities easily through the predicted result, it's a kind of symbolic information, then we can store an entity in these four aspects.

With this method, we actually store an entity in this pattern:

[Forward Context] [Entity Forward] [Entity Backward] [Backward Context]

As shown as \ref{candidate}, for every entity in ${\mathbf y^{i}}$, we can get four vector representations \{$\mathbf h_{b}^{f}, \mathbf h_{b}^{b}, \mathbf h_{e}^{f}, \mathbf h_{e}^{b}$\}, where the subscripts represent the beginning or end of the entity.  These four vectors have different meanings like we mentioned above, $\mathbf h_{b}^{f}$ has the context information before this entity, so we use it to represent forward context, while $\mathbf h_{e}^{b}$ is backward context, $\mathbf h_{b}^{b}$ and $\mathbf h_{e}^{f}$ are both entity information but one is from the entity beginning and the other one is from the end.  Then we easily concatenate these representations from all entities in the document to generate four matrices ${\{\mathbf r^{fc},\mathbf r^{eb},\mathbf r^{ee},\mathbf r^{bc}}\}$ as follows where the ${k}$ means ${k}$-th line of the matrix and also the ${k}$-th entity:
\begin{equation}
\label{E4}
\{\mathbf r_{k}^{fc},\mathbf r_{k}^{eb},\mathbf r_{k}^{ee},\mathbf r_{k}^{bc}\}=\{\mathbf h_{kb}^{f}, \mathbf h_{kb}^{b}, \mathbf h_{ke}^{f}, \mathbf h_{ke}^{b}\}
\end{equation}

In this example, both $\mathbf h_{t}^{f}$ and $\mathbf h_{t}^{b}$ of "Tao" contain this word itself, the $\mathbf h_{t}^{b}$ of "John" contains the information of "met", so we store them in the candidate pool to provide this decisive information for the reasoner to give out the inference result.  

\subsection{Reasoner and the Next Layer}

Based on the candidate pool, we actually store an entity as an object that has four describes just like our humans, in our human reading process, we make decisions refer to an entity recognized elsewhere because their description is similar.  So for every word to predict, we can use the similarity between the current word and the candidate pool from four aspects as a reference to make a better decision.  Every matrix in the candidate pool is actually a list of vector representations, which are also facts that contain a part of entities information, we can use a special multi-facts inference model to take suggestions from it.  

\begin{figure*}[h]
\includegraphics[scale=0.5]{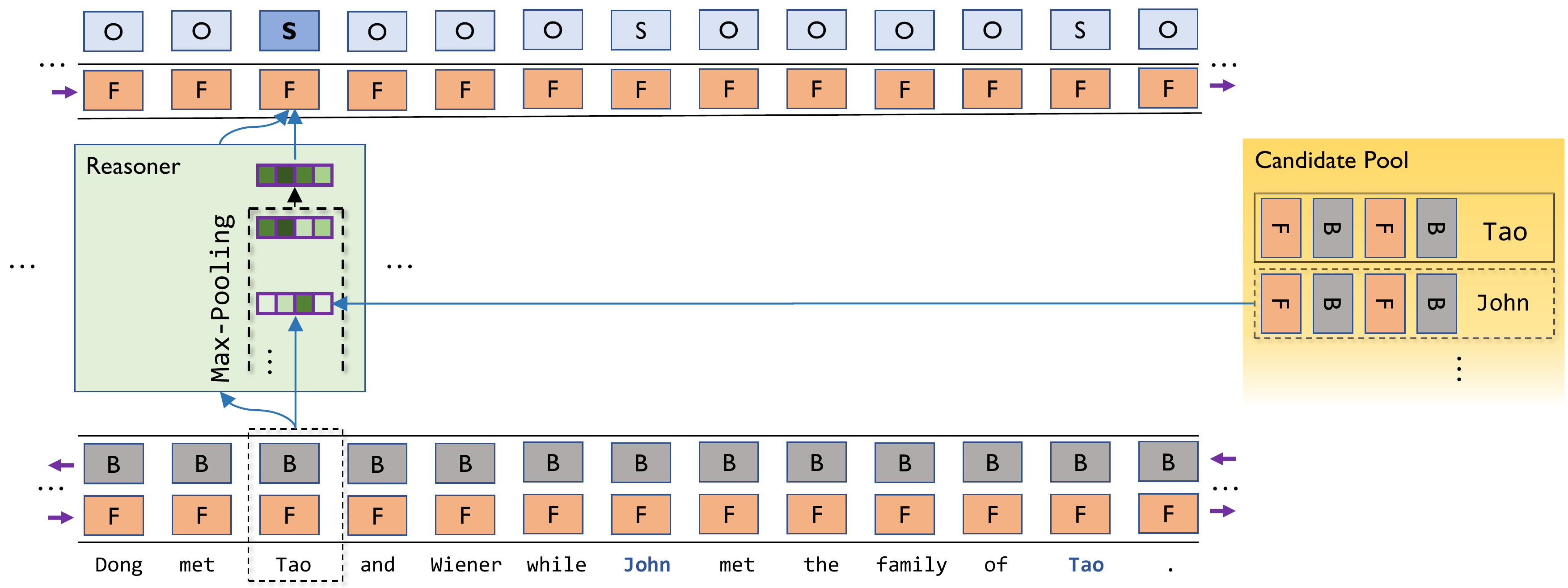}
\caption{The structure of reasoner which is also a multi-facts inference model. \label{reasoner}}
\end{figure*} 

As shown as \ref{reasoner}, the reasoner is actually a multi-facts inference model, in this model, the current word information is query, and the entities information in candidate pool is fact, we use a kernel $K(\mathbf query, \mathbf fact)$ to calculate the relations between the current word information and every entity information in candidate pool, the calculate results ${\mathbf s=\{s_{1},s_{2},s_{3},...,s_{n}}\}$ represent the suggestions from each entity recognized in the last layer.  Then we can get a final reference ${s}$ from these suggestions through the reasoner.  

In this work, we use similarity score to represent suggestions from entities and choose the dot product of vectors as the kernel.  Since we use the relations between words and entities in candidate pool to as an inference, the only important information is these relations themselves, but not which entity to give out this relation, so we don't need to take any content from the candidate pool to be involved in after calculation.  Furthermore, every calculate result in ${\mathbf s}$ represents the same type of relation, so we can use a pooling mechanism to get the most useful and typical value as the reference.

Like we did in \ref{memory}, we can divide the input of the decoder $\mathbf h_{t}$ into $\mathbf h_{t}^{f}$ and $\mathbf h_{t}^{b}$, and use these two vectors as query to get reference from the candidate pool respectively: 

\begin{equation}
\label{E5}
s_{t}^{fc}=max(sigmoid(\mathbf h_{t}^{f} \cdot {\mathbf r_{i}^{fc}})) \forall i\in [N_{e}]
\end{equation}
\begin{equation}
\label{E6}
s_{t}^{eb}=max(sigmoid(\mathbf h_{t}^{b} \cdot {\mathbf r_{i}^{eb}})) \forall i\in [N_{e}]
\end{equation}
\begin{equation}
\label{E7}
s_{t}^{ee}=max(sigmoid(\mathbf h_{t}^{f} \cdot {\mathbf r_{i}^{ee}})) \forall i\in [N_{e}]
\end{equation}
\begin{equation}
\label{E8}
s_{t}^{bc}=max(sigmoid(\mathbf h_{t}^{b} \cdot {\mathbf r_{i}^{bc}})) \forall i\in [N_{e}]
\end{equation}
where the $N_{e}$ means the number of entities.  

The calculating is just like a dot attention, the query is the input $\mathbf h_{b}^{b}$ and $\mathbf h_{e}^{f}$, and the key is the corresponding matrix, but we don't need any values because it doesn't matter which entity is similar to this input, just like we mentioned before, the real matter is whether there is a similar entity exists in NER task, so we just pick out the maximum value with max-pooling to represent the reference.

Then the vector ${\mathbf s_{t}=[s_{t}^{fc},s_{t}^{eb},s_{t}^{ee},s_{t}^{bc}]}$ can be viewed as a suggestion from the candidate pool, we feed this information to the decoder of next layer with ${\textsl H}$ together to make wiser decisions, the decoder shares the same parameters with the last layer, as follows:
\begin{equation}
\begin{split}
\label{E9}
\mathbf p_{t}^{i+1}&=Softmax(\textup{LSTM}(\mathbf h_{t},\mathbf s_{t},\mathbf p_{t-1}^{i+1}))\\
y_{t}^{i+1}&=argmax(\mathbf p_{t}^{i+1})
\end{split}
\end{equation}

Like the example in \ref{reasoner}, every entity stored in the candidate pool is composed of four vector representations come from the encoding information.  In the second layer, the decoder makes decision word by word, when it comes to the first "Tao", the $\mathbf h_{t}^{f}$ and $\mathbf h_{t}^{b}$ will get a high score through the computation with the representations of the last "Tao" in the candidate pool, then feed it to the decoder and recognize the first "Tao".  

\subsection{Training}
\label{training}
Since each NE-Reasoner layer has it's own predict results, which have a major influence on the next layer, so each layer needs to be trained by the supervision signals, and it's actually a co-training of multiple NER model.  If we use different encoder and decoder in different layers, then each layer could be trained independently as follows.  

\begin{equation}
\label{E10}
L(p(y|X))=-\sum_{i}\sum_{t}log(p(y^{i}|x_{t}))
\end{equation}

However, the NER models of each layer can share parameters most of the time like what we did in this work, which makes our model truly end-to-end.  In this situation, we can just train the last layer to learn to achieve reasoning more direct and have a good performance on all layers.  

\begin{equation}
\label{E11}
L(p(y|X))=-\sum_{t}log(p(y|x_{t}))
\end{equation}

\section{Experiments}

In this section, we will show the performance of our model on two different datasets, CoNLL-2003 English NER dataset, and a Chinese court judgment dataset. Since our model is a multi-layered structure model, the results of each layer are based on the refinement of last layer decoding results, so we will show the quantitative analysis like F1 score first, and then the qualitative analysis about how the final results outperform the first layer predict results on some specific cases.  

\begin{table}[h]
\begin{tabular}{l|l|l}
\hline
\textbf{Layer}          & \textbf{Parameter}  & \textbf{Size} \\ \hline
\multirow{2}{*}{CNN}    & character embedding & 30            \\ \cline{2-3} 
                        & window size         & 3             \\ \hline
Bi-LSTM                 & state size          & 256           \\ \hline
LSTM                    & state size          & 273           \\ \hline
NE-Reasoner Layers              & depth               & 2             \\ \hline
\multirow{2}{*}{Others} & batch size          & 16            \\ \cline{2-3} 
                        & learning rate       & 0.01          \\ \hline
                        & tag scheme          & BMEOS          \\ \hline
\end{tabular}
\caption{Hyper-parameters of NE-Reasoner for both experiments}
\label{hyper-parameters}
\end{table}

\subsection{CoNLL-2003 English NER}

The CoNLL-2003 dataset is a widely used NER dataset which has four types of named entities: person, location, organization and miscellaneous.  It has been separated into training, development and test sets.  

In order to be comparable with previous research results, we didn't preserve the document structure of sentences but shuffle them before training like other works, and we treat the mini-batch as a document.  In this work, we use the publicly available 50-dimensional word embeddings released by \cite{Pennington:14}.

\subsubsection{Result}

\begin{table}[]
\begin{tabular}{l|l|l}
\hline
\textbf{Char-Encoder}     & \textbf{Model}     & \textbf{F1}         \\ \hline
\multirow{5}{*}{None} & \cite{Collobert:11} & 88.67      \\ \cline{2-3} 
                      & \cite{Nguyen:16}    & 89.86      \\ \cline{2-3} 
                      & \cite{Huang:15}     & 90.10      \\ \cline{2-3} 
                      & \cite{Strubell:17}  & 90.54(±0.18) \\ \cline{2-3} 
                      & \textbf{NE-Reasoner}       & \textbf{90.78}      \\ \hline
LSTM                  & \cite{Lample:16}    & 90.94      \\ \hline
\multirow{3}{*}{CNN}  & \cite{Chiu:16}      & 90.91(±0.20) \\ \cline{2-3} 
                      & \cite{Ma:16}        & 91.21      \\ \cline{2-3} 
                      & \textbf{NE-Reasoner}       & \textbf{91.44}      \\ \hline
\end{tabular}
\caption{Evaluations on the test set of CoNLL-2003 English NER dataset}
\label{conll_score}
\end{table}

Some methods get high F1 score by introducing in extra features \cite{Chiu:16}, extra information from external datasets and pre-trained language model like \cite{Peters:17}, co-training language model like \cite{Liu:18}, or extra encoder on a different level of words like \cite{Zhang:18}.  NE-Reasoner could be implemented on these methods only if they are encoder-decoder based model.  Therefore, how NE-Reasoner achieve better performance than the baseline model in each layer by reasoning is more important.  

Since we used the CNN-Bi-LSTM-LSTM model as NER model of each layer, for a fair comparison, we compared with works that based on standard LSTM or CNN encoder without any external information resource.  

We implemented experiments with two settings for a better comparison: with and without the character level CNN.  As shown at \ref{conll_score}, our model can achieve better performance with both settings, which shows NE-Reasoner can work on different representations of words.  

\begin{table}[]
\begin{tabular}{l|l|l|l|l}
\hline
\textbf{Character Encoder} & \textbf{Layer} & \textbf{Precision} & \textbf{Recall} & \textbf{F1}    \\ \hline
\multirow{2}{*}{None}      & First          & 90.08              & 90.69           & 90.38          \\ \cline{2-5} 
                           & Final          & \textbf{90.33}     & \textbf{91.23}  & \textbf{90.78} \\ \hline
\multirow{2}{*}{CNN}       & First          & 90.24              & 91.95           & 91.09          \\ \cline{2-5} 
                           & Final          & \textbf{90.45}     & \textbf{92.45}  & \textbf{91.44} \\ \hline
\end{tabular}
\caption{Difference of each layer in NE-Reasoner on the test set of CoNLL-2003 English NER dataset}
\label{NE-Reasoner_diff}
\end{table}

\ref{NE-Reasoner_diff} shows the F1 score gets an obvious improvement after refinement from the first layer, and the most important reason is the second decoding layer results can get a better recall rate, which is in line with the expectation of model design.  

The two results of the first layer are comparable with other works, that means the training of our model won't hurt the first decoding layer results, and the improvement is totally from the correction of NE-Reasoner.   

\subsubsection{Analysis}

\begin{table*}[htb]
\centering
\begin{tabular}{l|p{12cm}}
\hline
\textbf{Decoder} & \textbf{Result}                                                                                    \\ \hline
first layer result       & The London club had been rocked by a two-goal burst from forwards \underline{Dean}(PER) Sturridge and \underline{Darryl Powell}(PER) \\ \hline
final layer result      & The London club had been rocked by a two-goal burst from forwards \underline{Dean Sturridge}(PER) and \underline{Darryl Powell}(PER) \\ \hline
first layer result       & \underline{Hiroshige Yanagimoto}(PER) cross towards the goal which \underline{Salem Bitar}(ORG) appeared to have covered             \\ \hline
final layer result      & \underline{Hiroshige Yanagimoto}(PER) cross towards the goal which \underline{Salem Bitar}(PER) appeared to have covered             \\ \hline
\end{tabular}
\caption{Typical cases of CoNLL-2003 English dataset to show how the second decoding layer refine the results}
\label{conll_case}
\end{table*}

\begin{table*}[htb]
\centering
\begin{tabular}{l|p{12cm}}
\hline
\textbf{Decoder} & \textbf{Result}                                                                                    \\ \hline
first layer result      & \begin{tabular}[c]{@{}l@{}}被告人\underline{光头强}(PER)窜至某宿舍...\\ The accused \underline{Guang Touqiang} (PER) sprang to a domitary...\\光头强趁机作案...\\
Guang Touqiang took the opportunity to commit\end{tabular}                                                                             \\ \hline
final layer result      & \begin{tabular}[c]{@{}l@{}}被告人\underline{光头强}(PER)窜至某宿舍...\\ The accused \underline{Guang Touqiang}(PER) sprang to a domitary...\\ \underline{光头强}(PER)趁机作案...\\
\underline{Guang Touqiang} (PER) took the opportunity to commit\end{tabular}                                                                             \\ \hline
first layer result      & \begin{tabular}[c]{@{}l@{}}被告人偷取\underline{李某甲}(PER)现金2000元...\\ The accused stole 2000 yuan from \underline{Li Moujia}(PER)\\被告人偷取常士禄现金3000元...\\The accused stole 3000 yuan from Chang Shilu\end{tabular}                                                                                                                                                      \\ \hline
final layer result     & \begin{tabular}[c]{@{}l@{}}被告人偷取\underline{李某甲}(PER)现金2000元...\\ The accused stole 2000 yuan from \underline{Li Moujia}(PER)\\被告人偷取\underline{常士禄}(PER)现金3000元...\\The accused stole 3000 yuan from \underline{Chang Shilu} (PER)\end{tabular}                                                                                                                                                      \\ \hline
\end{tabular}
\caption{Typical cases of chinese court judgement to show how the second decoding layer refine the results}
\label{judgment_case}
\end{table*}

We choose some typical cases from the test dataset to explain how our model works.  As shown at \ref{conll_case}, the model can refer to the entities recognized elsewhere. In the first example, the model can realize that "Sturridge" is also a part of a name cause it saw the another recognized name "Darryl Powell" in the second layer.
In the second example, the model predicted "Salem Bitar" as an organization at first but then changed it to a person with the help of reasoner.  

\subsection{Chinese Court Judgment NER}

The Chinese court judgment is a sub-task of a court judgment document parsing task, and also an application of NER model on specific fields.  The context patterns and the entities are usually mentioned multiple times in a judgment, so the document structure is important to help us to dig out the missed entities in the first decoding layer results.  

\subsubsection{Result}

\begin{table}[H]
\centering
\setlength{\tabcolsep}{1mm}{
\begin{tabular}{|l|l|l|l|}
\hline
\textbf{Model}         & \textbf{Pcs} & \textbf{Recall} & \textbf{F1} \\ \hline
first layer result     & \textbf{97.26}        & 96.13           & 96.87       \\ \hline
final layer result       & 97.21        & \textbf{97.26}           & \textbf{97.23}       \\ \hline
\end{tabular}}
\caption{Evaluations on the test set of our Chinese court judgment dataset}
\label{judgment_score}
\end{table}

There is no obvious hierarchical structure in Chinese compared with English because Chinese is composed of Chinese characters directly, so the CNNs character encoder in English is useless.  The other hyper-parameters are the same with \ref{hyper-parameters}.  

As the same as the result in CoNLL-2003 English dataset, NE-Reasoner also achieved a better F1 score on Chinese NER task with the improvement of recall rate, as shown at \ref{judgment_score}.  The precision rate decreased because NE-Reasoner might predict some false positives through the reasoner.  

\subsubsection{Analysis}

As shown at \ref{judgment_case}, each case contains two sentences which come from the same document, the model process these two sentences separately, but the NE-Reasoner still works because the candidate pool is built on the entire document.  In the first case, the Chinese character "光"(Guang) is not a common last name in training dataset, so the model missed it in the first layer, except the first mention because the context pattern is too strong, then the reasoner gives out a high similarity score to help to recognize it successfully in the second layer.

In the second case, the missed entity in the first layer is just mentioned once, but the context pattern is exactly the same with another entity in the same document, then the similarity score helps the model find it in the second layer.

The only difference between the layers is the reference from reasoner, and to make sure the result is convincible, we replaced the name at the first mention with another name and then the second layer just keep the first results, we also changed the similarity score manually and got the same result, so it's definitely the role of the reasoner.  

\section{Conclusions and Future work}

In this work, we proposed a NE-Reasoner to mimic the human reading process and keep global consistency in the NER task.  NE-Reasoner can learn to integrate the local decisions in the same document and make the inference, experiments show that our model can learn to extrapolate from the recognized entities information to make wiser decisions so as to achieve better performance.  

The method to store information to candidate pool and make the inference from that through reasoner is symbolic and quite effective, which uses label information to select the content of entities information and just calculate a similarity score but not a representation through reasoner.  

However, the similarity score calculated by the kernel is not always reliable, like some words are totally different but the similarity score is quite high, because neural networks don't have enough knowledge to really distinguish these words like us humans, so we will study how to get a more accurate and more efficient kernel.  

NE-Reasoner is not only useful in NER task, the fusion of local decisions and inference from reasoner could be used on many tasks, we can also apply our model to other NLP tasks like POS tagging or chunking etc, and even on some sequence generating tasks.


\bibliographystyle{aaai}
\bibliography{aaai19}






\end{CJK}
\end{document}